\documentclass{llncs}
\usepackage{float}
\usepackage{tabularx,ragged2e}
\usepackage{amsmath}
\usepackage{booktabs}
\usepackage{multirow}
\usepackage{subfigure}
\usepackage{lscape}
\usepackage[table,xcdraw]{xcolor}
\usepackage{makeidx}  
 \usepackage{multirow}
 \usepackage[table,xcdraw]{xcolor}
\usepackage{amssymb}
\usepackage{array}
\newcolumntype{P}[1]{>{\centering\arraybackslash}p{#1}}
\usepackage{graphicx}
\usepackage{tabu}
\usepackage[flushleft]{threeparttable}
\usepackage{url}
\usepackage{algorithm}
\usepackage[noend]{algpseudocode}
\makeindex
\begin{document}

\title{Diabetic Foot Ulcer Grand Challenge 2022 Summary}

\author{Connah Kendrick\inst{1}\orcidID{0000-0002-3623-6598} 
	\and Bill Cassidy \inst{1}\orcidID{0000-0003-3741-8120} 
	\and Neil D. Reeves \inst{2}\orcidID{0000-0001-9213-4580} 
	\and Joseph M. Pappachan \inst{3}\orcidID{0000-0003-0886-5255} 
	\and Claire O'Shea \inst{4} 
	\and Vishnu Chandrabalan \inst{3}\orcidID{0000-0002-2687-1096} 
	\and Moi Hoon Yap\inst{1}\orcidID{0000-0001-7681-4287}}
%
\authorrunning{C. Kendrick et al.}
%

\institute{Department of Computing and Mathematics, Manchester Metropolitan University, Manchester M1 5GD, UK \and
	Musculoskeletal Science and Sports Medicine, Manchester Metropolitan University, Manchester M1 5GD, UK \and
	Lancashire Teaching Hospitals NHS Foundation Trust, Preston, PR2 9HT, UK \and
	Waikato District Health Board, Hamilton 3240, New Zealand \\
	\email{Connah.Kendrick@mmu.ac.uk}
}

\maketitle


\begin{abstract}
	The Diabetic Foot Ulcer Challenge 2022 focused on the task of diabetic foot ulcer segmentation, based on the work completed in previous DFU challenges. The challenge provided 4000 images of full-view foot ulcer images together with corresponding delineation of ulcer regions. This paper provides an overview of the challenge, a summary of the methods proposed by the challenge participants, the results obtained from each technique, and a comparison of the challenge results. The best-performing network was a modified HarDNet-MSEG, with a Dice score of 0.7287.
\end{abstract}

\section{Introduction}

Initial works \cite{goyal2020recognition,Goyal2017b} in diabetic foot ulcer (DFU) analysis demonstrated that deep neural networks had the potential to give reliable classification and segmentation outcomes for DFU regions. However, these works were limited by small dataset size (705 images), low image resolution ($500 \times 500$ pixels) and non-clinical annotations. In more recent works, Yap et al. \cite{yap2020deep,yap2020diabetic} hosted the first DFU detection and classification challenges, in conjunction with MICCAI 2020 and MICCAI 2021. Then, followed by Wang et al. \cite{wang2020fully,rostami2021multiclass} who hosted an online MICCAI 2021 event in The Foot Ulcer Segmentation Challenge (FUSeg). They combined a series of datasets from the AZH wound care centre and released a 2020 dataset containing 1109 $224 \times 224$ images. Then, they released a newer dataset \cite{rostami2021multiclass} for the FUSeg challenge, containing 1210 images, with a resolution of $512 \times 512$ pixels. In addition, 160 images from the Medetec wound dataset was included, resized to $560 \times 391$ pixels. The data was annotated by clinical staff at the AZH foot clinic, but was low resolution and included black borders to pad the images. 

\begin{figure}[!htb]
	\centering
	\includegraphics[width=1.0\textwidth]{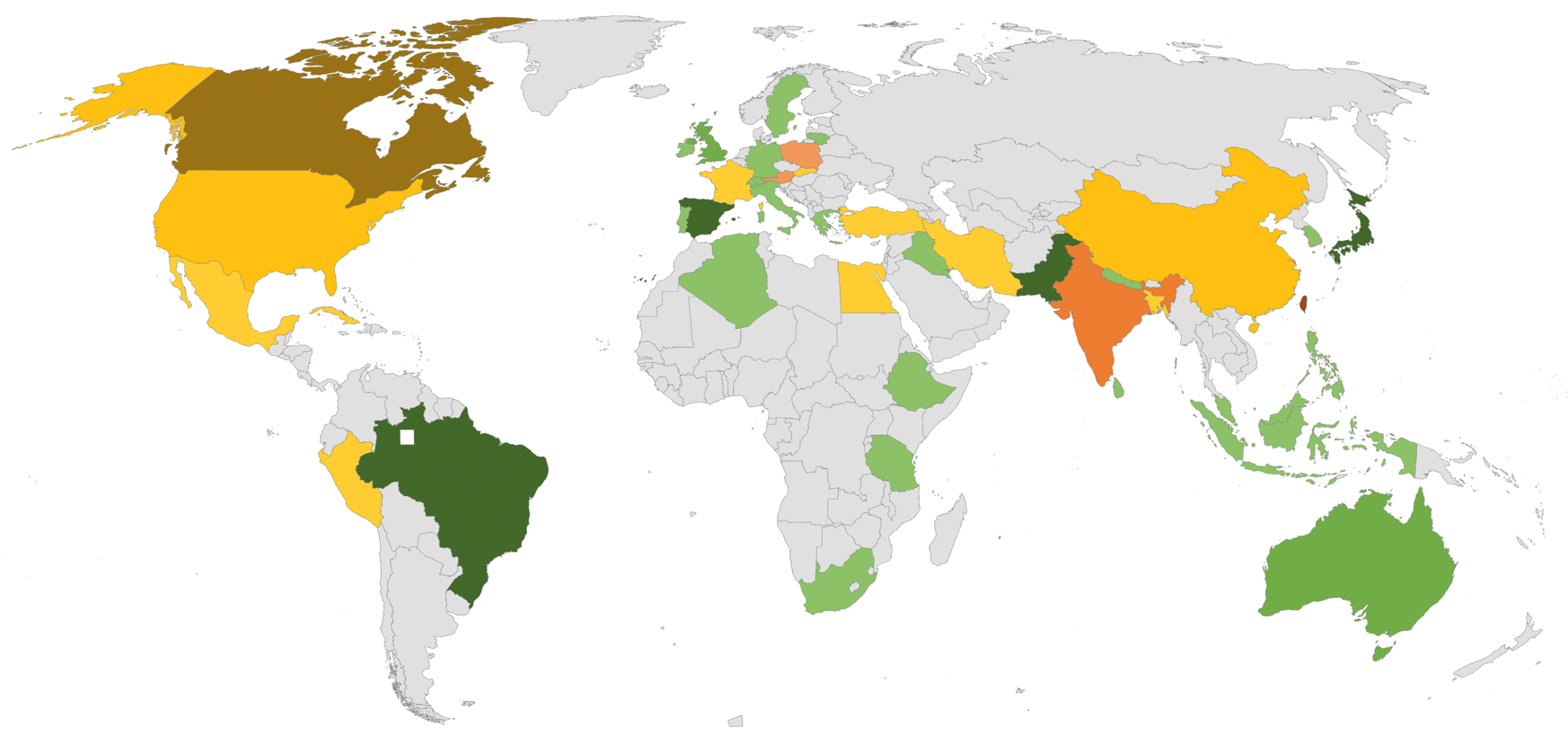}
	\caption{Distribution of researchers by country of origin who used the DFUC2022 datasets.}
	\label{figure:dfuc2022_users}
\end{figure}

In 2022, Kendrick et al. \cite{kendrick2022translating} introduced the DFUC2022 dataset, which contains 2000 high resolution DFU images with clinical annotations, and 2000 images without annotations for use as the test set. When compared to last year's challenge \cite{yap2020diabetic}, the total number of participating countries increased from 25 to 47. Figure \ref{figure:dfuc2022_users} highlights the distribution of the DFUC2022 dataset users.

\section{Methodology}
This section summarises the creation of the DFUC2022 dataset, the performance metrics, and presents an analysis of the methods proposed by the participants of the challenge.

\subsection{Datasets and Ground Truth}

The DFUC2022 dataset is the largest publicly available DFU segmentation dataset and has the highest resolution images. The dataset consists of 4000 images collected from diabetic foot clinics at the Lancashire Teaching Hospitals NHS Foundation Trust, UK. Each image was collected using one of three cameras by medical photographers during patient appointments. During each appointment, medical photographers took photographs of the ulcers at a distance of approximately 30-40cm without the aid of a tripod \cite{yap2021development,joseph2022future}. In some cases, the photographs show a consistent background (blue or white), but others showed no background. These issues highlight the challenges inherent in manual medical photography in clinical settings.

The dataset was annotated by 5 podiatrists using the VGG annotator \cite{dutta2016via,dutta2019vgg} software to create polygon outlines of the DFU wound regions. The dataset was then processed using an active contour algorithm \cite{kroon2022snake} which smoothed the manually delineated regions that were drawn around the DFU regions. For further details, please refer to Kendrick et al. \cite{kendrick2022translating}.

For DFUC2022, we split the dataset into two sets of 2000 images, one for training, and the second for validation and testing. The training set comprises a variety of DFU cases, where individual feet may exhibit multiple DFUs, resulting in a total of 2304 ulcers in the training set. The dataset comprised cases of the same wound at different stages of development, whereby cases may be healing or may have becoming more severe. Due to the whole foot nature of the images and some early cases, a large portion (89\%) of the DFU wounds were less than 5\% of the total image size. However, the dataset still comprised DFU within the size range between 0.04\% - 35.04\% of the total image size.

\subsection{Performance Metrics} 

We compared the performance of the deep learning networks on the accuracy of the segmentation using Dice coefficient, Jaccard, False Positive Error (FPE) and False Negative Error (FNE). We used image-based metrics so that multiple lesions in a single image were treated as one whole lesion. 
To determine the leaderboard ranking we used the Dice coefficient \ref{eq:DICE}. The equation is based on Intersection over Union (IoU), but doubles the amount of pixels. This provides the metric with preserved sensitivity and resistance to outliers. In addition to providing a value between $0 - 1$, where $0$ is no overlap and $1$ is an exact overlap.

\begin{equation}
Dice = 2 * \frac{|X \cap Y|}{|X| + |Y| - |X \cap Y|}
\label{eq:DICE}
\end{equation}

We also included the Jaccard index, also called IoU. The metric compares the area of overlap over the area of union. It also outputs values similar to Dice.

\begin{equation}
IoU = \frac{|X \cap Y|}{|X| + |Y|}
\label{eq:JACCARD}
\end{equation}

The FPE indicates the percentage in which a selected system falsely predicts a non-DFU pixel as DFU. 
\begin{equation}
FPE = \frac{FP}{FP + TN}
\label{eq:FPE}
\end{equation}

The FPE indicates the percentage in which a selected system falsely predicts a DFU pixel as non-DFU.
\begin{equation}
FNE = \frac{FN}{FN + TP}
\label{eq:FNE}
\end{equation}

For both the Dice and the Jaccard metrics, the metrics assume overlap has occurred, if no overlap is present we score 0 for that image.

\subsection{Summary of the Proposed Methods}
In this section, we summarise the methods used by the top 10 entries for the DFUC2022. We present the methods in reverse order, i.e., start from the 10th position and through the leaderboard top rank. We provide a brief overview of the methods used and the scores that each achieved.

The method that ranked 10th in the challenge was submitted by AGH\_MVG (the team from AGH University of Science and Technology, Poland). They used an end-to-end ensemble of 3 models in their proposed pipeline, namely, YOLOv4 \cite{shinde2018yolo} to localise the DFU, and a Vision Transformer DETR \cite{xu2021co} with a U-Net module to segment the detected DFU regions. They used an augmentation technique to train the model in which the extracted DFU region has different percentages of visibility allowing the network to learn from semi-occluded images. Their approach achieved a Dice score of $0.6725$.

The method which placed 9th was submitted by IISLab (Technical University of Kosice, Kosice, Slovakia and Department of Burns and Reconstructive Surgery, PJ Safarik University and Hospital Agel, Kosice-Saca, Slovakia). They modified the nnU-Net \cite{isensee2019nnu} architecture to work with 2D data rather than 3D data that the model had originally been designed for. They implemented two versions of the model, with and without skip connections to bridge data between convolutional and deconvolutional layers of the U-Net. Additionally, they analysed the use of cutmix \cite{yun2019cutmix} and mixup \cite{zhang2018mixup} augmentations and observed that using a combination of both increased the performance of the network. Overall, their best model achieved a Dice score of $0.6975$.

The 8th place method was submitted by DGUT-XP (Dongguan University Of Technology, China). They achieved a Dice score of $0.6894$. However, they did not submit their method description for inclusion into the summary paper.

The method in 7th place was submitted by GP\_2022 (Alexandria University, Egypt and University of Southern California, USA). They used an ensemble of two models: DeepLabV3+ \cite{chen2018deeplab} and SegFormer \cite{xie2021segformer}. They used pretraining for both models with the ade20k dataset \cite{zhou2017scene}, but with different optimisers. They applied standard augmentation techniques to the images during training, and ensembled the results from both networks to produce the final prediction. Additionally, they included a post-processing stage for hole filling and the removal of small pixel regions. They achieved a Dice score of $0.6986$.

The method in 6th place was submitted by FDHO (University of Applied Sciences and Arts Dortmund, Germany). They used a combined-refinement approach where they generated realistic DFU images using StyleGan+ADA \cite{karras2020training}. They also trained a series of models based on a Feature Pyramid Network using the modified ResNeXt-101 backbone to include the squeeze and excite layer. The models were used to annotate the generated images, which were used to train a series of extended models under 5-fold cross validation. To aid the process, they performed data augmentation over the dataset. In addition, they used post-processing to remove small pixels regions in the prediction results. They achieved a FID score of $19.09$ on the generated images, demonstrating the ability to generate a diverse dataset as well as a Dice score of $0.7169$ for the ensemble of extended models.

The method which placed 5th was submitted by Seoyoung (Sogang University, South Korea). They used a Swim Transformer based backbone for the upper network. They padded the dataset by using a series of augmentation techniques resulting in a total of 8000 training images which included the original training images. They also experimented using DeepLabV3+, OCRNet, and ConvNeXt \cite{liu2022convnet}. Using their approach, they achieved a Dice score of $0.7220$, however, the author did not submit a full paper.

The method in 4th place was submitted by Adar-Lab (National Yang Ming Chiao Tung University, Taiwan). They implemented a two-stage approach, firstly they used Fast R-CNN \cite{girshick2015fast} to perform object detection on the high-resolution image. From the Fast R-CNN model they extract regions of interest, then they used TransFuse \cite{zhang2021transfuse} for segmentation. In their experiment, they use different fusion techniques to merge transformer and CNN outputs. They used a unique augmentation approach for the model architecture, where for every original image, they cropped 4 images from each corner, with the size of $384 \times 384$ pixels. In their training, they used a combined loss of pixel-based Binary Cross-Entropy (BCE) and IoU. They achieved a Dice score of $0.7254$.

The method placed 3rd was submitted by agaldran (Universitat Pompeu Fabra, Spain). They implemented a two-length cascade of encoder decoder networks, the encoders were pretrained on imagenet, whereas the decoders used feature-pyramid networks \cite{lin2017feature}, which improved the feature extraction of DFU at different sizes. Their experiments showed that the ResNeXt101 model had the best performance. They further increased the performance by performing a study of loss metrics, i.e., BCE training only, Dice training only, BCE with slow interpolation to Dice, BCE and a hard switch to Dice, and BCE+Dice training. They showed that for DFU and polyp segmentation the combined BCE+Dice work performed best and provided valuable insights into the effect of these combined losses on the training process. They achieved a Dice score of $0.7263$.

The method which placed 2nd was submitted by LkRobotAILab (West China Hospital, China). Their method is based on the state-of-the-art segmentation network OCRNet \cite{liu2021polarized}. They completed an ablation study using an alternative model backbone that uses a mix of convolutional and transformer-based structures. Their study showed improvements to network performance when using ConvNeXt-XLarge with pre-training on ImageNet-21K. The dataset images were processed using gamma correction to help standardise during inference. Finally, to refine network performance, they used boundary loss which focuses on the edges of the predicted segmentation regions. They achieved a Dice score of $0.7280$.

The winning entry for DFUC2022 was submitted by yllab (National Tsing Hua University, Taiwan). Their method was inspired by the state-of-the-art in polyp segmentation in colonoscopy videos. The network was based on HarDNet-MSEG \cite{Huang2021}, which uses an encoder-decoder structure. The encoder improved upon the Densenet \cite{huang2017densely} structure by reducing the number of shortcuts and increasing channel count. The decoder takes input from the encoder at multiple stages through specialised Receptive Field Block (RFB) modules which are then aggregated via dense layers and upsampled for the final segmentation result. They improved the HarDNet-MSEG model by balancing the inputs and outputs of the encoder by splitting the channels of the input, as well as adjusting the skip connections to different layers. The decoder is then replaced with a Lawin Transformer \cite{yan2022lawin}. They then use post processing to fill holes in the resulting segmentation map. Using this method, they achieved the best Dice score of $0.7287$. 

\section{Results and Discussion}
In total, we received 1320 submissions throughout the whole challenge process. In addition, a total of 22 teams participated in the challenge, with 50 individual users registered on the DFUC2022 Grand Challenge website. There were approximately 1100 submissions for the validation stage. During the validation stage, we allowed a maximum of 10 submissions per-day with a maximum of 200 prediction masks per submission. The best performing team for the validation stage was LkRobotAILab, with a Dice score of $0.7156$, a Jaccard score of $0.6195$, and a FNE score of $0.2352$. We note that team seoyoung achieved the best FPE score of $0.2312$. 

During the release of the test set, we observed that the positions on the leaderboard changed. Table \ref{table:baselines} compares the performance of the Top-10 submissions. Yllab was the winning team, with a Dice score of $0.7287$, an improvement of $0.1579$ over the baseline scores of the test set (0.5708), provided by the organisers \cite{kendrick2022translating}. LkRobotAILab scored the best Jaccard score of $0.6276$, with $0.1727$ improvement over the baseline scores. Agadran scored $0.2210$ on FPE, with $0.1614$ improvement over the baseline. Finally, Adar-Lab scored $0.1847$ for FNE, with $0.0654$ improvement over the baseline result. 

The results demonstrate an improvement over the baseline scores in most cases. This highlights the ability of the proposed techniques to segment DFU regions. However, in the case of FNE, we observed only a minor improvement on the scores. This is most likely due to the ratio of DFU pixels in the images, compared to the non-DFU pixels. As highlighted by some participants, duplication identification, where the same image had been labelled by more than one clinician, demonstrated disagreement between annotators which could potentially improve the performance of segmentation if resolved. After the challenge, the organisers have opened a live leaderboard, where at the time of writing this paper, the best performance is reported by Kendrick et al. \cite{kendrick2022translating}, with a Dice score of 0.7446. 

\begin{table*}[!h]
	\centering
	\renewcommand{\arraystretch}{1.0}
	\caption{The top-10 participating teams in DFUC2022, starting with the best Dice score.$\dagger$ = higher score is better; $\uplus$ = lower score is better. \textbf{Bold} indicates the best overall result.}
	\label{table:baselines}
	\begin{tabular}{|l|c|c|c|c|}
		\hline 
		Team &\multicolumn{4}{c|}{Metrics} \\
		
		&Dice $\dagger$ & Jaccard $\dagger$ & FPE $\uplus$ & FNE $\uplus$ \\
		\hline
		\hline
		yllab & \textbf{0.7287}&0.6252&0.2048&0.2341\\
		\hline
		LkRobotAILab& 0.7280&\textbf{0.6276}&0.2154&0.2261\\
		\hline
		Agadran& 0.7263&0.6273&0.2262&\textbf{0.2210}\\
		\hline
		Adar-Lab & 0.7254&0.6245&\textbf{0.1847}&0.2582\\
		\hline
		seoyoung & 0.7220&0.6208&0.1925&0.2584\\
		\hline
		FHDO & 0.7169&0.6130&0.214&0.2453\\
		\hline
		GP\_2022& 0.6986&0.5921&0.2065&0.2778\\
		\hline
		DGUT-XP & 0.6984&0.5945&0.2523&0.2379\\
		\hline
		IISlab & 0.6974&0.5926&0.2163&0.2734\\
		\hline
		AGH\_MVG& 0.6725&0.5690&0.2555&0.2830\\
		\hline
	\end{tabular}
\end{table*}




\section{Conclusion} 

In this study, we introduced the largest publicly available DFU segmentation dataset, annotated by clinical experts working in diabetic foot clinics. We propose a supervised technique to track the wound progression and recovery at point of care. This dataset is not the first publicly available DFU segmentation dataset, however, our dataset contains the largest number of images at a higher resolution than those datasets previously released. We provide this new dataset to the research community to encourage further research and progress in the field. The advancements shown in our proposed method have the potential to be used to support clinical staff in hospital settings and to assist with at-home tracking of DFU by patients. Thus, aiding the treatment process and preventing future complications such as infection and resulting limb amputation. 

The networks demonstrated in this paper show that although deep learning can provide accurate results, there is still much work to do in the task of automated segmentation of DFU wounds. These challenges include the ability to segment early onset and small DFU regions in whole foot images, the differential between DFU and peri-wound regions, and contextual wound information such as signs of healing and degradation.

This work can help to progress the field by using such models as part of fully automated DFU monitoring systems that can be used at home or in foot clinics to track DFU healing progress and to relieve burden on healthcare workers. This work will build on our existing framework \cite{cassidy2021cloud} in delivering an easy-to-use system capable of advanced forms of diabetic foot analysis, which will include longitudinal monitoring as a means of assessing wound healing progress.

\section*{Acknowledgment}
We would like to thank the MICCAI conference for hosting DFUC2022, and AITIS for sponsoring the wining teams' prizes. We would also like to thank all participants of the challenge for their effort and contributions to DFU research.

\addtocmark[2]{Author Index} 
\renewcommand{\indexname}{Author Index}
\printindex

\bibliographystyle{unsrt}
\bibliography{Ref}

\end{document}